\documentclass[sigconf,nonacm]{acmart}
\usepackage{subcaption}
\usepackage{dirtree}

\AtBeginDocument{%
  \providecommand\BibTeX{{%
    \normalfont B\kern-0.5em{\scshape i\kern-0.25em b}\kern-0.8em\TeX}}}

\begin{document}

\title{An Embedded and Real-Time Pupil Detection Pipeline}

\author{Ankur Raj}
\affiliation{%
  \institution{Ubiquitous Computing\\University of Siegen}
  \streetaddress{Hoelderlinstr. 3}
  \city{Siegen}
  \country{Germany}
  \postcode{57076}
}
\author{Diwas Bhattarai}
\orcid{1234-5678-9012}
\affiliation{%
  \institution{Ubiquitous Computing\\ University of Siegen}
  \streetaddress{Hoelderlinstr. 3}
  \city{Siegen}
  \country{Germany}
  \postcode{57076}
}

\author{Kristof Van Laerhoven}
\affiliation{%
  \institution{Ubiquitous Computing\\University of Siegen}
  \streetaddress{Hoelderlinstr. 3}
  \city{Siegen}
  \country{Germany}
  \postcode{57076}
}

\renewcommand{\shortauthors}{Raj, Diwas, and Van Laerhoven}

\begin{abstract}
Wearable pupil detection systems often separate the analysis of the captured wearer's eye images for wirelessly-tethered back-end systems. We argue in this paper that investigating hardware-software co-designs would bring along opportunities to make such systems smaller and more efficient. 
We introduce an open-source embedded system for wearable, non-invasive pupil detection in real-time, on the wearable, embedded platform itself. 
Our system consists of a head-mounted eye tracker prototype, which combines two miniature camera systems with Raspberry Pi-based embedded system. Apart from the hardware design, we also contribute a pupil detection pipeline that operates using edge analysis, natively on the embedded system at 30fps and run-time of 54ms at 480x640 and 23ms at 240x320. Average cumulative error of 5.3368px is found on the LPW dataset for a detection rate of 51.9\% with our detection pipeline. For evaluation on our hardware-specific camera frames, we also contribute a dataset of 35000 images, from 20 participants.
\end{abstract}
\maketitle

%%%%%%%%%%%%%%%%%%%%%%%%%%%%%%%%%%%%%%%%%%%%%%%%%%%%
\section{Introduction}
%%%%%%%%%%%%%%%%%%%%%%%%%%%%%%%%%%%%%%%%%%%%%%%%%%%%
Eye tracking applications have generated several commercial products \cite{pupilOpenSource,TobiiProLab}, initially for complementing user surveys or improving designs in shopping centres, products and advertisement placements \cite{cognetive_process_eye_tracking}. In automobiles it has similarly been used to measure drivers' fatigue \cite{kapitaniak_walczak_kosobudzki_jóźwiak_bortkiewicz_2015_automobile_accident,jamil_mohammed_awadalla_2016_sleep_automobile}. 
Eye tracking is an especially important modality, as it has become  ubiquitous in commercial products and serves many applications: it has been used to study users' visual attention \cite{Borji2013StateoftheArtIV_attention}, sleep detection and emotional state, and shows immense promise in Augmented Reality (AR) and Virtual Reality (VR) applications for foveated rendering \cite{foveated_tracking_nvidia}, i.e., increasing the amount of detail in images based on the user's focus or fixation points. 
It includes behavioural information like visual-attention \cite{visual_attention_aviation,Borys2017EyetrackingMI_visual_attention}, emotional state \cite{lim_mountstephens_teo_2020_emotion} as well as medical information. 
Eye tracking products also increasingly become smaller and more efficient, as they are integrated in mobile and wearable devices as an additional biometric, for instance for unlocking a smartphone \cite{patel_han_jain_2016_smartphone_eye}.
Crucial to any eye tracking device is the pupil detection step, which is needed before performing other routines such as eye tracking, gaze estimation, and developing interfaces based on eye movement.

\begin{figure}[b]
    \centering
    \includegraphics[width=\columnwidth]{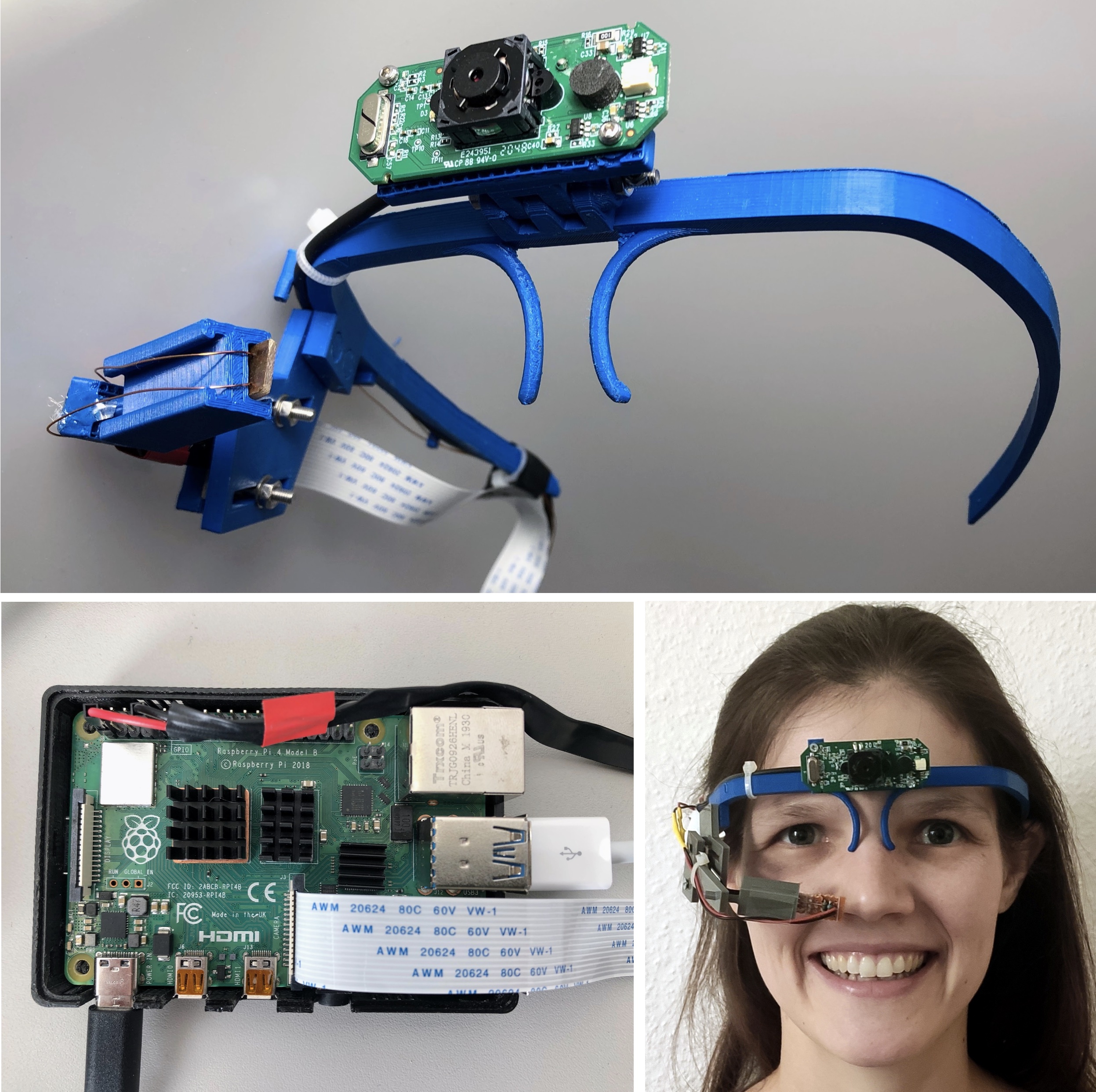}\vspace{-2mm}
    \caption{The 3D-printed hardware setup connects two cameras through USB and CSI, along a LED power delivery using GPIO to the Raspberry Pi Model 4 B. The bottom-right image shows the front view of the hardware while worn. }
    \Description{}
    \label{fig:hardware_setup_and_subject_wearing}
\end{figure}

Pupil detection has been researched extensively, leading to many commercial eye trackers being available \cite{TobiiProLab,pupilOpenSource} nowadays. These provide their end users with APIs (Application Programming Interface) for collection of data and performing eye and gaze tracking. Such eye trackers tend to be expensive and platform-specific, focusing strongly on high-end camera features to cater for all potential users. 
We argue in this paper that for many applications, the resulting high-speed and high-resolution cameras might prevent a light-weight and energy-efficient pupil detection pipeline. For such cases, we propose an approach that uses a miniature and low-cost hardware design, which can be replicated by others with minimal effort. For the software design, we developed an embedded and real-time pupil detection pipeline based on edge analysis, and contribute a dataset of near eye images taken from this hardware for validation. We thus present a comprehensive hardware-software open-source framework that allows researchers to holistically design their own pupil tracker.
The major contributions of our paper can thus be summarized as follows:
\begin{itemize}
    \item We present an embedded hardware design for embedded eye tracking, based around commercial and widely available camera modules and a Raspberry Pi platform. 
    \item We developed an open-source pupil detection pipeline that can run natively on the Raspberry Pi, in real-time.
    \item We created a benchmark dataset of near-eye images with our hardware design, from people of different ethnicity, gender, eye colour and age. 
    \item We evaluate our pipeline on this dataset and a public one \cite{DBLP:journals/corr/TonsenZSB15} for performance in detection speed and accuracy on an embedded platform. 
\end{itemize}

The remainder of this paper is structured as follows:
In section \ref{sec:related_work}, the state-of-the-art in this field is described, where we focus particularly on existing detection methods and different hardware setups, as well as on state-of-the-art datasets used in eye tracking. We position our contributions among these.  
In the following section \ref{sec:methods}, the design methodology for the hardware and the detection method that we designed to work on the embedded system are presented, as well as our strategy for the creation of a near eye benchmark dataset. This is followed by the evaluation (section \ref{sec:evaluation}), which contains the validation of our detection pipeline on the proposed hardware, and also contains the performance of the pupil detection on an existing state-of-the-art public dataset. We finally conclude this paper in section \ref{sec:conclusions} where we discuss the main results and give an outlook on this research.

%%%%%%%%%%%%%%%%%%%%%%%%%%%%%%%%%%%%%%%%%%%%%%%%%%%%
\section{Related Work} \label{sec:related_work}
%%%%%%%%%%%%%%%%%%%%%%%%%%%%%%%%%%%%%%%%%%%%%%%%%%%%

Eye tracking and pupil detection depends on various components. In this section, we introduce the closest areas in eye tracking research, particularly the hardware elements that capture images of the eye, the detection pipeline to analyze those images, and available datasets that are used to measure the performances of the detection strategies. 
Several pupil detection and gaze estimation methods have been introduced over these past years and numerous researchers have used different methodologies, ranging from blob detection \cite{set_javadi_hakimi_barati_walsh_tcheang_2015}, edge analysis \cite{pupilOpenSource, ExCuSe2015,santini_fuhl_kasneci_2018_PuRe} to machine learning \cite{NV_Gaze_kim_stengel_majercik_de,kothari2020ellseg, pupil_net_2016} based approaches. 
We focus here on pupil detection as a first step towards creating an eye tracking and gaze estimation setup. Prior research has focused on developing a fast and accurate pupil detection pipeline, that could also work in a variety of challenging scenarios. 

\subsection{Eye Tracking Hardware Designs} \label{subsec:related_work_hardware}

Any pupil detection strategies are highly dependent on the hardware being used for capturing high resolution images of the eye, especially in different lighting conditions, frame rate, and position of the camera. Due to these parameters, the final detection method heavily relies on the selection of the hardware setup. Such setups that are commonly used for pupil detection can be divided into two sub-categories; (1) \emph{Remote Eye Setup} and (2) \emph{Near Eye Setups} or \emph{Head Mounted Camera} systems. 

Remote Eye Hardware setups have the camera situated away from the participant. It can follow both a multiple camera \cite{beymer_flickner_2003,ohno_mukawa_2004_remote_multiple} or a single camera \cite{remote_eye_hosp_eivazi_maurer_fuhl_geisler_kasneci_2020,meyer_böhme_martinetz_barth_2006_remote_single} design. 
Free movement of the user's head is allowed within a fixed range. 
The camera images of the eye, head and posture of the whole body is captured with these systems. 
They provide high resolution cameras with high frame rates (>500 Hz), with some commercial eye tracking systems even reaching frame rates of more than 1000 Hz \cite{tobiiScreenMultiple}. These high frame rates are not usually observed with the near eye setups.
The remote setups are useful in making inferences about the behaviour of the user, like visual-attention \cite{Borji2013StateoftheArtIV_attention}, drowsiness, and fatigue \cite{rodrigue_son_giesbrecht_turk_höllerer_2015,driving_hu_lodewijks_2021,fatigue_8464177}, which could be essential in sending alerts if an automobile driver becomes drowsy \cite{driving_hu_lodewijks_2021}. 
Hardware designed by researchers includes custom remote eye tracking setups such as \cite{remote_eye_hosp_eivazi_maurer_fuhl_geisler_kasneci_2020, meyer_böhme_martinetz_barth_2006_remote_single,hennessey_noureddin_lawrence_2006_remote}, mostly due to its own feature requirements and the expensive commercial alternatives. 
A recent open-source setup is RemoteEye \cite{remote_eye_hosp_eivazi_maurer_fuhl_geisler_kasneci_2020}, which provides a high-speed eye tracker that could achieve frame rates of more than 500 \emph{fps} \cite{remote_eye_hosp_eivazi_maurer_fuhl_geisler_kasneci_2020} during pupil detection, and provides a gaze estimation accuracy of 0.98 degrees, on average \cite{remote_eye_hosp_eivazi_maurer_fuhl_geisler_kasneci_2020}. It provides a low-cost alternative to commercial high speed remote camera systems.

Near Eye Hardware Setups or head mounted camera system uses an eye tracker headset with one (monocular) or two (binocular) eye cameras and/or a world camera. It can be enclosed e.g. VR setups or can be open e.g. Google Glass or Microsoft Holo lens. Additionally, it has numerous infrared LEDs to provide consistent lighting and contrast between pupil and iris. The setups also use USB or other wired/wireless connectivity. The near eye tracker is mobile and could capture higher detail images of the eye. 
Researches have made their own near eye trackers from commercial components. The authors \cite{eye_tracker_2004} modified safety glasses to attach analogue cameras and IR LEDs to create an invasive eye tracking system, where the point of gaze is estimated offline. openEyes \cite{li_babcock_parkhurst_2006_open_eyes} is an open source eye tracking contribution that uses two IEEE-1394 web cameras (640x480 @ 30 Hz) for synchronised frames and Starburst detection algorithm. The authors \cite{ryan_duchowski_birchfield_2008_LIBMUS_Eye_tracker} have extended the framework with the ability to seamlessly switch from watching the pupil/iris border in bright light to tracking the iris/sclera boundary (limbus) in low light. Researches have disassembled web camera like Microsoft LifeCam VX-1000 to create a low-cost DIY eye tracker \cite{mantiuk_kowalik_nowosielski_bazyluk_2012_DIY_eye_tracker}. The Newer eye trackers include frames that are 3D printed \cite{lukander_jagadeesan_chi_müller_2013_omg,borsato_technology_morimoto_paulo_2019_stroboscopic} and use stroboscopic lighting \cite{borsato_technology_morimoto_paulo_2019_stroboscopic}. Furthermore, research has also been performed on a low-cost hardware to produce a smart eye tracking system \cite{smart_eye_tracker_raspberry_pi_wheelchair_8369701}, that connects with a wheelchair to assist elderly people in gaining mobility via smart commands. Another study implementing an embedded system (Raspberry Pi) in eye tracking is from the author(s) \cite{ferhat_vilarino_sanchez_2014_raspberry_pi}, where the open-source gaze tracking software (\emph{openGazer}) \cite{zieliński_2013_open_gazer} has been extended upon and tested for evaluation on an embedded device. 
Other well-known commercial products for near eye tracking are from Pupil Labs \cite{pupilOpenSource, pupil_net_2016} and Tobiipro \cite{TobiiProLab}. 

As suggested by the authors of \cite{remote_eye_hosp_eivazi_maurer_fuhl_geisler_kasneci_2020}, one disadvantage of the near eye system is lower frame rates (30Hz - 120Hz), which hinders in the application of robust saccade and micro-saccade detection. So far, only limited research exist in using a head mounted, multi-camera eye tracking system incorporated within a fully embedded system, whereas this field is particularly interesting because it would grant more mobility to the end user, as the device would be portable enough that it can be carried along or be worn by the user.

\subsection{Pupil detection methods} \label{chp:related_work_detection_methods}

Historically, a majority of pupil detection methods employ a range of well-known computer vision techniques. They predominantly rely on edge detection methods, thresholding, and morphological filtering to locate the pupil in the current image. Recently, more end-to-end machine learning based approaches have gained in popularity and are also being used for pupil detection. Any of these methods have to face similar challenges while performing pupil detection, however, with pupil detection being mostly hampered by challenges that include:
\begin{itemize}
    \item reflection in eye images due to external lighting
    \item blury images due saccade and micro-saccade
    \item poor contrast between iris and pupil
    \item obstruction/occlusion in pupil view due to eyebrows, eyelid and/or eye wear.
\end{itemize}

A well known computer vision based pupil detection pipeline is proposed by Lech \'Swirski et al. \cite{Swirski2012Bulling}. In this pipeline, the input image is first convolved with a HAAR-like feature to have a maximum response within the image. This convolution operation is repeated for different user-defined radii sizes. The radii with maximum response is used as a pupil region. The pupil is identified in this region by k-means clustering on the histogram, followed by edge detection using canny filter to finalize the boundary. Lastly, a Random Sample Consensus (RANSAC) ellipse fitting is used to detect the pupil. PupilLabs \cite{pupilOpenSource} follow an approach for pupil detection that is similar to the pipeline used in article \cite{Swirski2012Bulling}. After ROI estimation and edge detection, filtering of edges is performed based on histogram analysis and removal of any spectral reflection. The contours are then connected, and ellipse fitting is performed on the identified pupil.

In another study \cite{set_javadi_hakimi_barati_walsh_tcheang_2015} a detector using threshold operation has been proposed. After the identification of pupil segments, a convex hull operation is used to enclose the segments and the enclosure is used to identify the centre of pupil. ExCuSe \cite{ExCuSe2015} is yet another well known state-of-the-art method for pupil detection in real world scenarios. It uses two different approaches to find the pupil, being edge detection and thresholding. If there is a peak in the bright region of the histogram analysis, detection can be performed using edge analysis \cite{ExCuSe2015, fuhl_tonsen_bulling_kasneci_2016}.
PuRe (or Pupil Reconstructor) is also an edge-based detector that uses edge segment selection and conditional segment combination schemes to perform pupil detection \cite{santini_fuhl_kasneci_2018_PuRe}.  Additionally, it provides a confidence measure ($\psi$) for the selected ellipse.

Machine learning based methods are known to be used in the research of pupil detection and gaze estimation. The ML based estimators use features for training a neural network model to locate the pupil and determine the gaze. Convolutional Neural Network (CNN) and VGG are the commonly used ML models.
A CNN based ML model is NVGaze \cite{NV_Gaze_kim_stengel_majercik_de}, it provides accurate pupil detection and gaze estimation on synthetic as well as real near eye images. It ihas little modifications, depending on the application. The model includes 6 - 7 convolutional layers each with 2×2 stride, and dropout layers after each convolutional layer, and no padding or pooling. 
PupilNet is another well known CNN based detector \cite{pupil_net_2016}. It uses \emph{dualCNN} pipeline, where one network coarse estimates the location of pupil and the other network fine positions on this estimate to have a final localisation of pupil. Estimation for occluded eye images due to eyelids, wearables or even reflections make the pupil detection a difficult task. EllSeg framework \cite{kothari2020ellseg} also offers a method to segment the entire elliptical structure containing the iris and pupil, hence providing pupil detection on highly occluded images. The EllSeg framework can be used with any encoder-decoder architecture, as well. 

\subsection{Datasets: Pupil \& Gaze Estimation} \label{sec:datasets}

With the rise in augmented reality and virtual reality applications and machine learning (ML)-based estimators, there are now more publicly available datasets, containing images or videos of Near Infrared (NIR) \cite{DBLP:journals/corr/TonsenZSB15}, remote eye images \cite{cvpr2016_gazecapture} and simulated eye models \cite{NV_Gaze_kim_stengel_majercik_de}, with manual annotation of pupil position. Table \ref{table:dataset_comparison} contains some of the most well-known state-of-the-art datasets for pupil detection and gaze estimation to provide an overview of their 
 features.

\begin{table*}[htb]
    \centering
    \caption{State-of-the-art publicly available datasets, the lighting conditions within these datasets can be varied i.e. indoors and outdoors. For article \cite{open_eds_meta} a controlled environment is created to capture the images inside a VR headset.}

    \resizebox{\textwidth}{!}{%
    \begin{tabular}{p{0.15\textwidth} p{0.2\textwidth} p{0.2\textwidth} p{0.3\textwidth} c c} 
     \toprule
     \textbf{Dataset} & \textbf{Number of images} & \textbf{Lighting conditions} & \textbf{Hardware used} & \textbf{YOP} & \textbf{Gaze Direction}  \\ [0.5ex] 
     \midrule
     GazeCaptue  \cite{cvpr2016_gazecapture} & 2.5M & Indoors and outdoors & Mobile/tablet camera & 2016 & yes\\ 
     
     NVGaze \cite{NV_Gaze_kim_stengel_majercik_de} & 2.5M real and 2M  synthetic  & Indoors with varied lighting & VR headset and a head mounted camera 640x480 @ 30 Hz and 1280x960 & 2019 & yes\\ 
     
     Labelled Pupils in the wild \cite{DBLP:journals/corr/TonsenZSB15} & 130,856 & Indoors and outdoors & Head mounted camera; 480x640 @ 95 Hz & 2016 & no\\
     
     OpenEDS \cite{open_eds_meta} & 356,649 & Indoors, in a controlled environment & VR headset; 400x640 @ 200 Hz & 2019 & no \\
     
    \bottomrule
    \end{tabular}
    }
    
    \label{table:dataset_comparison}
\end{table*}

LPW or Labelled pupil in the wild \cite{DBLP:journals/corr/TonsenZSB15} contains 66 high quality videos from 22 participants, where the videos were recorded using a head mounted camera system \cite{pupilOpenSource}. The dataset is collected under varying lighting conditions, both indoors and outdoors, at 95$fps$. The pupil location is manually labelled inside the dataset. Gaze estimation is not part of the dataset, but it contains pupil centre location annotation for each frame.

NVGaze \cite{NV_Gaze_kim_stengel_majercik_de} contains near eye images in infrared lighting. It has both synthetically generated data (2M images at 1280×960) and real-world data (2.5M images at 640x480) collected from 30 subjects. Synthetic data was rendered from 3D models \cite{wood_baltruaitis_zhang_sugano_robinson_bulling_2015} under active illumination with 4 IR LEDs, with each image labelled with 3D eye location, 2D pupil location and 2D gaze vector. The synthetic dataset contains exact segmentation for sclera, skin, pupil and iris. The real-world data consists of binocular images captured from a 120Hz camera. The data is collected under two separate conditions, with users performing acuity tasks. One using VR hardware with IR illumination, and the other using a head mounted setup for AR with varying infrared intensity, emulating a real world scenario with varied lighting conditions.

GazeCapture is a dataset that is used for gaze estimation \cite{cvpr2016_gazecapture}. It is a crowd-sourced project where users capture their images using mobile devices like phones and tablets. The subjects use the mobile application to capture images of their face while they look at markers on the screen. The images are captured in a variety of lighting conditions, making it a diverse dataset. There are more than 2.5M images in the dataset. Since the images are taken in different lighting conditions and from a remote device (mobile phone or tablet), near eye images of the pupil are not part of this dataset. The dataset also contains motion data from the mobile device (accelerometer, gyroscope, manometer) captured at 60Hz.  

Open EDS \cite{open_eds_meta} contains near eye infrared images in a controlled environment. The images were captured using a VR headset that has been modified to capture images in 200 Hz. The dataset contains binocular images with a subset of images that have pixelated annotation of eye regions (eyelid, pupil, iris). These annotations were made using ellipse selector as well as manual annotators. Since this dataset contains not only the centre of pupil but also annotation for each region of eye on real-world participants, it becomes a novel dataset for training ML and non-ML based detectors.

\section{Methodology} \label{sec:methods}

The eye tracking hardware is the central component of this paper, which provides near eye infrared images in high quality that would be used for our pupil detection pipeline. It is important to have a low-cost and open-source hardware platform that makes it accessible for any researchers to incorporate and eye tracking in their projects. We have created a custom monocular hardware design that has an eye camera and a world camera. The work from the pupil labs \cite{pupilOpenSource} and previous eye trackers \cite{li_babcock_parkhurst_2006_open_eyes,eye_tracker_2004} served as an inspiration for this hardware design.

The eye tracker can be divided into 3 major components, the frame, the camera systems \& their connection, and the compute-device. Except the compute device, all the components are attached to the eye tracker frame.
% frame weight
The frame of the eye tracker draws inspiration from safety eyewear and DIY eye tracker from Pupil Labs, and can be worn as an accessory. The eye tracker headset/frame is designed in a 3D modelling software and is 3D printed using PLA as a print material. Individual camera support holders are printed separately and joined together via snap connectors. The eye tracker is monocular and the frame incorporates a holder for the world camera (fig \ref{fig:hardware_setup_and_subject_wearing}).

The camera system can be further subdivided into the eye camera and world camera. The eye camera is a Raspberry Pi Zero camera with an OmniVision OV5647 camera sensor from Sparkfun. It is a fixed focus 5 MP camera that supports  VGA(640x480) @ 90Hz and 720 p @ 60 Hz. The focus of the camera is manually modified to have it capture near images of the eye. The eye camera (Pi Zero NoIR) supports taking images in infrared lighting, as it is build without the infrared blocking filter and an infrared LED array is created with SFH 4050-Z from OSRAM Opto Semiconductors. %The emitter allows one to take images of the eye with a consistent lighting condition. 
The world camera (Logitech HD-C615) takes images of the subject's Field of View (FOV), this commercial camera needs to be disassembled first to be attached with our headset. It supports 1080p/30fps and 720p/30fps resolutions. 

The central processing module is a Raspberry Pi model 4B, a miniature and low-cost single-board computer made by from Raspberry Pi foundation that runs on a Debian based Linux distribution (Pi OS bullseye).
This board is used to perform pupil detection locally, on the . It has an ARM Cortex A72 64-bit SoC @ 1.5GHz and 4 GB of RAM. The compute-device provides the necessary interface to our cameras and LED array. The eye camera is connected to a camera adapter to be later connected to the compute device with a CSI cable, the world camera is connected with a USB cable to the compute-device, and the IR-emitter is controlled with GPIO connections to the Raspberry Pi. 

\begin{table*}[ht]
  \caption{List of components in the Eye Tracker with all the component being available commercially, where the headset is created using a custom design, that could accommodate the camera system. The 3D model design of the headset can be found in this repository : \url{https://github.com/anonymized.edu/pupiltracker}}
  \label{tab:cost_breakdown}
  \begin{tabular}{ccl}
    \toprule
    Components & Product Name & Specifications\\
    \midrule
    Eye Camera  & Pi Zero Camera NoIR  &  5MP VGA(640x480) @ 90Hz and 720 p @ 60 H\\
    World Camera & Logitech HD-C615 & 1080p/30fps and 720p/30fp UVC complient\\
    Compute Embedded Device  & Raspberry Pi model 4 B & ARM Cortex A72 64-bit SoC @ 1.5GHz, 4 GB RAM\\
    Headset & Custom 3D printed design & \\
    Emitter IR-LEDs & OSRAM SFH 4050-Z & \\
  \bottomrule
\end{tabular}
\end{table*}

\begin{figure*}[htb]
    \centering
    \includegraphics[width=\textwidth]{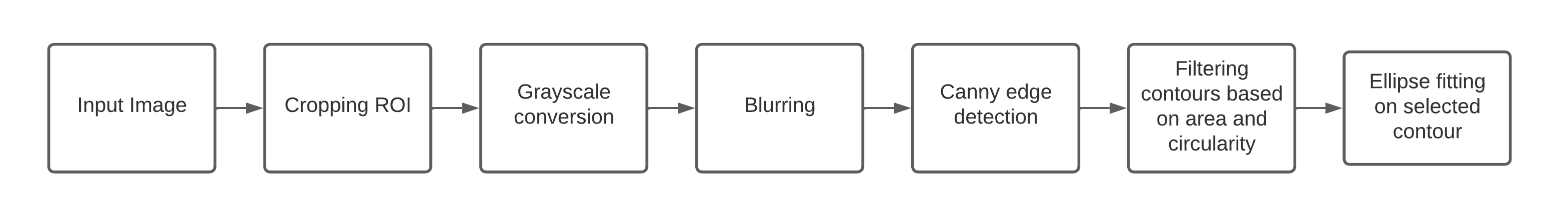}
    \caption{Our pupil detection pipeline, which is implemented in Python3 and makes extensive use of the OpenCV library \cite{opencv_library}, is made available open-source. It uses ellipse fitting, where the input image is capture in infrared lighting, and confidence on the parameter selection for canny ($T_{canny}$) and blurring ($K_{blur}$) is performed using the parameter estimation routine (sec \ref{sec:parameter_estimation_pupil}).}
    \Description{Pupil detection pipeline shown with various components in blocks}
    \label{fig:pipeline_pupil_detection}
\end{figure*}

The hardware is used to acquire the near eye images in IR lighting, with the initial constraint of having a detection method to be able to run on our embedded device. We wanted our detection strategy to have as few components as possible, such that it would not be computationally expensive and would not stall the processor of the Raspberry Pi. This motivated us to create a custom ellipse fitting method, which is not computationally expensive and also provides detection of pupil in a variety of edge cases. The ellipse fitting methods use edge detection to find contours in the image, and from these contours the contour representing the pupil is identified and segmented. Figure \ref{fig:pipeline_pupil_detection} shows our pipeline components, which is similar to one presented by the authors of \cite{ExCuSe2015} and relies on edge features of the images to find the pupil in the initial image. The pupil is assumed to be dark and have a distinguishing boundary as compared to the iris when illuminated by the infrared LED lighting.

%% Pipeline pupil detection figure

\paragraph{Preprocessing} 
The initial image is captured in infrared lighting of resolution 480x640 or 240x320 px. This image is cropped for the Region of Interest (ROI), and contains the region of eye including the eyebrows and eyelashes. The ROI cropped image is converted to greyscale and is blurred using a median blur filter with a fixed kernel size ($K_{blur}$). Canny edge detection with intensity threshold ($T_{canny}$) is performed to obtain the edges of the image. A morphological open operation is used to reduce the intensity of the eyelashes \cite{Swirski2012Bulling}.

It is important to note at this poit that the choice of the canny edge detector intensity ($T_{canny}$) and median blurring ($K_{blur}$) could lead to images with various goodness in edge response. It is therefore crucial to find a set of $T_{canny}$ \& $K_{blur}$ values that would produce the best identifiable edges for pupil. This is investigated in section \ref{sec:evaluation}. Once we have an image with clearly identifiable features for pupil, morphological operations are performed to find which edges could correspond to pupil.

\paragraph{Filtering contours for pupil detection}

\begin{enumerate}
    \item After the determination of initial edges in the frame with preprocessing, the contours inside the frame are identified and filtered to correspond to the pupil. For this, an estimation of the closed area of the contour is made using convex hull operation. This ensures that, only close contours are considered as a prospective pupil, hence, straight line edges are discarded. %% could write an expression for it
    %%%%A% Expression for Convex hull operarion open cv
    
    \item  The area of the closed contour is used as a filtering strategy, since the pupil size in a frame is usually found to be constant or lie in a range (\emph{MAX\_PUPIL\_SIZE} \& \emph{MIN\_PUPIL\_SIZE}). Based on this, we obtain a list of closed contours that could correspond to pupil.
    
    \item Filtering is then performed based on circularity of the closed contour. The eye-camera is in front of the subject's eye, leading to the pupil appearing circular. Therefore, we can filter the contours which do not lie above a circularity value ($T_{circularity}$). The circularity is calculated by formula \ref{chp:methods_circularity_formula}.
    
      \begin{equation}
          \label{chp:methods_circularity_formula}
        circularity\_hull = \frac{(4 * \pi * area\_hull)}{circumference\_hull ^ 2}
     \end{equation}
     
\end{enumerate}

With these operations a closed contour is identified, that lies within the estimated area of pupil, and is above a circularity threshold value. Now an ellipse can be fitted onto the identified closed contour. Estimation of the centre of the pupil can be made on the selected contour. 

%%%%%%%%%%%%%%%%%%%%%%%%%%%%%%%%%%%%%%%%%%%%%
%%%%%%%%%%%%%%%%% Dataset %%%%%%%%%%%%%%%%%%%
%%%%%%%%%%%%%%%%%%%%%%%%%%%%%%%%%%%%%%%%%%%%%

The third component of a detection system is the dataset of images, which can be used for analysis of different detection methods and can be tested with different computer architectures. It is common in the research area of eye tracking to create a dataset specific to the hardware configuration \cite{NV_Gaze_kim_stengel_majercik_de, DBLP:journals/corr/TonsenZSB15}. Since our DIY hardware is made from off-the-shelf components, there is no publicly available dataset that is specific to our camera selection. The images are obtained in two different resolutions, i.e. 480x640 and 240x320 px. 
The lower resolution images are useful when we would like to have a faster detection system, as it reduces the IO operation time. The higher resolution images are useful when we would want to have more details on our images, i.e. edges are more prominent. All the images for the dataset are taken in indoor conditions with an open window and/or incandescent lighting in the room. The display/monitor is also a source of light during this experiment. The setup represents a normal indoor condition where there could be light coming from different light sources, instead of a controlled environment. 

Two different approaches are taken for data collection. First, a \emph{fixed marker location} where positional markers are added to a screen, the participants are asked to look at fixed positional markers on the display. A constant distance of 60 cm from the screen is maintained while performing the experiment. Second, \emph{free movement} where the participants are asked to %look in every direction randomly, 
perform a clockwise/counterclockwise movement of the eye, maintain gaze towards different objects. The participants maintained a fixed head pose while they performed both the experiments. 

\begin{figure}[htb]
  \centering
  \includegraphics[trim=0 120 0 0, clip, width=0.99\linewidth]{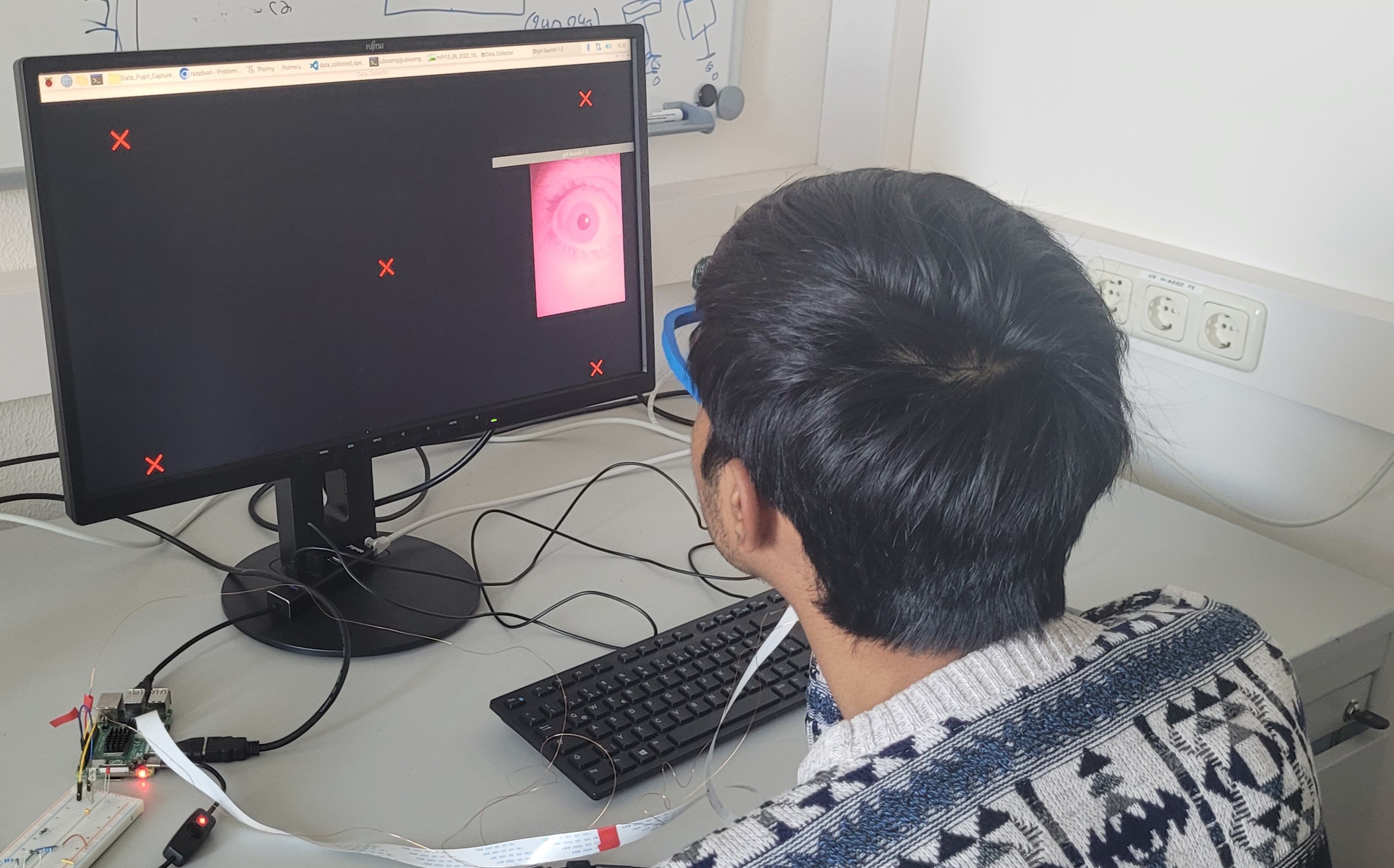}
  \caption{Study participants perform the experiments, while these run completely on the embedded system itself. The Raspberry Pi (left) acquires the images, while in the fixed marker location experiment each participant would start from the top left marker location, and would go to top right marker, followed by central marker, then bottom left marker, and finally bottom right marker. Displayed here are all markers and the real-time eye camera feed, which  is not displayed to the participant while performing the experiment.}
  \Description{A male sitting infront of a computer screen and looking at positional markers while wearing the hardware. The compute device can be seen on the left side of the image, with cable connections to the eye tracker}
  \label{fig:subject_fixed_marker_location}
\end{figure}

In Dataset Marker Location, fixed positional markers are displayed on a monitor of resolution 1920x1080 px. The participants are asked to look at these marker locations sequentially (figure \ref{fig:subject_fixed_marker_location}).
For each marker location, 30 images of the participant's eye are captured. Since the participants are told to look at the marker locations, there are only a few blinks while performing the experiment. And the participant maintained a constant gaze. In total 150 images are captured per session of the data collection. The participants performed one experiment for the higher resolution (480x640 px), and then repeated the same procedure for a lower resolution (240x320 px). 

Dataset Free Movement is the second data collection procedure, the participants are not asked to look at fixed markers, but instead, they are encouraged to look in different directions while maintaining a fixed head position. The participants rolled their eyes in clockwise and counterclockwise, moved their eye randomly, and sometimes even fixated on different objects in the room. The aim of this method is to get more edge case data, where the participants are not bound to a fixed location marker or even keep their eye open, during the entire length of the experiment. The data collection is also performed for two different resolutions (640x480 and 240x320 px), and 300 images of the eye are taken in each session. The \emph{free\_movement} procedure provided images where there is blinking, occluded eyes and blurry frames. These images are helpful in performance evaluation of the pupil detection pipeline.

%%%%%%%%%%%%%%%%%%%%%%%%%%%%%%%%%%%%%%%%%%%%%%%%%%%%
\section{Evaluation}\label{sec:evaluation}
%%%%%%%%%%%%%%%%%%%%%%%%%%%%%%%%%%%%%%%%%%%%%%%%%%%%

The ellipse detection pipeline is evaluated with the LPW dataset \cite{DBLP:journals/corr/TonsenZSB15}, it provides near eye infrared images of 22 different participants. The image repository of each participant contains near eye images (2000 frames) in different lighting conditions, which are called parts in the dataset. We use the term "use case" to represent the parts in a participant repository, for example, use case \emph{LPW/P1/P9} represents Participant 1 Part 9 of the dataset. Since the LPW dataset has a lot of challenging images e.g. outdoors, participants with prescription glasses, reflections and shadows, it is therefore selected as a benchmark for testing our pupil detection pipeline.

\subsection{Model Evaluation} \label{subsec:accuracy}

% how we measure the accuracy
The Euclidean distance or L2 norm is used as a method for testing the accuracy of the model. The error between the evaluated centre of pupil from our detector and the actual centre of pupil is calculated for every participant in the dataset. The initial frame from each participant (480x640 px) is cropped for region of interest (ROI) and the default processing is used. Changes are only made to the $T_{canny}$, $K_{blur}$ and the size of pupil to perform detection. 

\begin{figure}[t]
    \centering
    \begin{subfigure}[t]{0.7\columnwidth}
        \centering
        \includegraphics[height=1.6in]{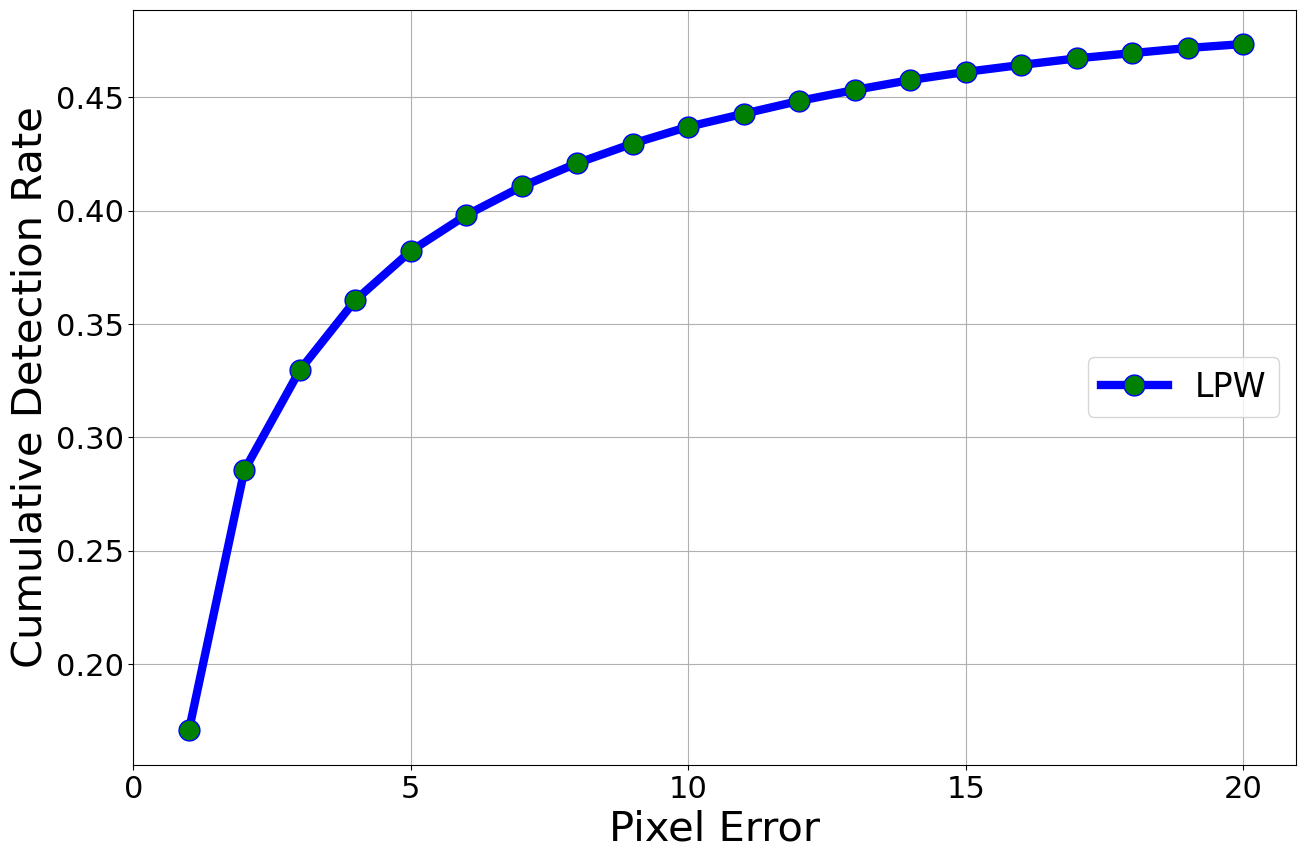}
    \end{subfigure}%
    ~
    \begin{subfigure}[t]{0.2\columnwidth}
        \centering
        \includegraphics[height=1.6in]{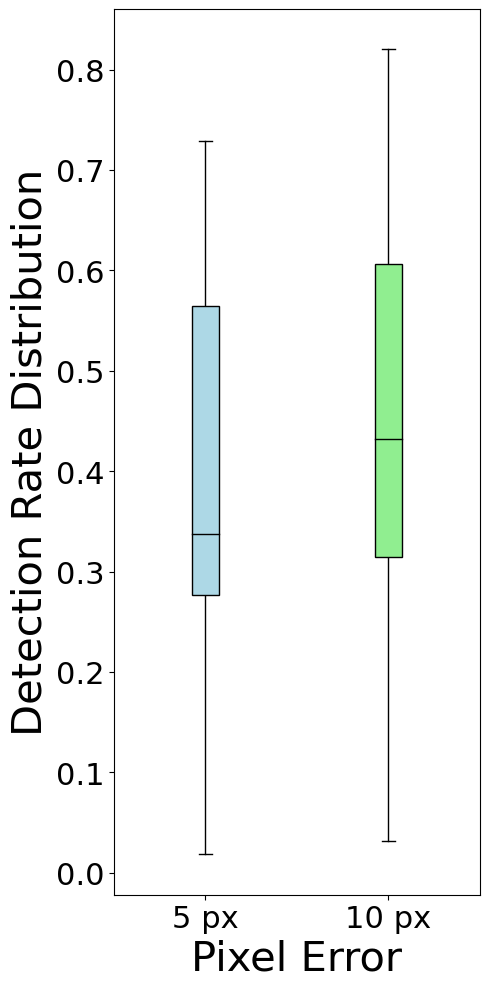}
    \end{subfigure}
    \caption{Cumulative error distribution for our ellipse detection method on the LPW dataset \cite{DBLP:journals/corr/TonsenZSB15} up-to 20px error. The detection rate distribution at 5px and 10px is shown in the right.}
    \Description{Two plots are presented to display the accuracy of our detection method on LPW dataset. The left figure is a line graph with pixel error on the x-axis and cumulative accuracy on the y-axis. This plot starts from 10\% accuracy and goes till 48\% accuracy for 20 px error. The other figure (box plot) represents the error distribution at 5 and 10 pixels. The lowest detection rate at 5 pixels is less than 10\% and highest is 75\%. For 10 px the lowest distribution is again less than 10\% and highest is 85\%.}
    \label{fig:box_plot_destribution_accuracy}
\end{figure}

Our ellipse detection pipeline performed at a cumulative accuracy of 38.21\% with less than 5px error and  43.6\% with less than 10px error and an overall average accuracy of 51.9\% at 5.3368 px. The detection rate is comparable to \emph{Pupil Labs}, \emph{ExCuSe} and \emph{\'Swirski} of 30\% detection rate with lower than 5px error on LPW dataset \cite{DBLP:journals/corr/TonsenZSB15}. Our detection method stagnates after 20px error at close to 50\% accuracy on the LPW dataset. However, the current ellipse detector is able to perform near accurate detection on complicated dataset without using any extensive feature extraction method e.g. HAAR or ML based operation. The LPW dataset contains challenging images that have high reflections, strong shadows and eyelid occlusions \cite{DBLP:journals/corr/TonsenZSB15}. Images in outdoor scenarios can be challenging, especially in case the participant is wearing prescriptive glasses, resulting in strong reflections. This is one of the reason for the lower detection distribution of less than 10\% in these extreme cases (fig \ref{fig:box_plot_destribution_accuracy}). One example of this case is with \emph{LPW/P5/P6}, where the pupil is obstructed with eye wear, another example is  \emph{LPW/P20/P7} where the pupil is not in focus resulting in weak edges for pupil. For some use cases, as suggested in \emph{PuRe} \cite{santini_fuhl_kasneci_2018_PuRe}, detection accuracy can be improved by readjustment of the position of eye camera.

\subsection{Parameter Estimation} \label{sec:parameter_estimation_pupil}

As discussed before, the embedded edge detection pipeline, works on pre-determined set of parameters for Canny threshold ($T_{canny}$ ) and Median blur kernel size ($K_{blur}$). Since there are multiple combinations of $T_{canny}$ \& $K_{blur}$ that could be used for pupil detection, it is important to  find the best parameter set that could provide detection for most cases. To achieve this goal, a parameter estimation routine is created that uses a variety of near eye images to produce the best estimate on these parameters ($T_{canny}$ \& $K_{blur}$).

\subsubsection{Procedure for parameter estimation}

All near eye infrared images of each participant are stored in a separate directory. Only images that have a visible pupil are stored in the working directory. Edge cases that have less than 50\% of the visible pupil, closed eye and/or high reflections are omitted, since they could affect the parameter estimation result, and there might be no pupil detection. The objective of the procedure is to have images where the pupil detector is bound to work, and if there is no definitive pupil in the image, there would not be any detection, which could add a penalty to the loss parameter. Next, using a list of threshold and kernel size values, the routine takes a combination of $T_i$ \& $K_i$ values and uses it to perform pupil detection on all the images in the working directory. The program then calculates a loss for each image or frame in the directory, this gives an estimate on how good the pupil detection was for this combination. The loss parameter $L$ is evaluated as :-

%%%A%% Loss function
\begin{equation}
    \label{chp:evaluation:loss_formula}
    L = N_{c} + A_{min} + C
\end{equation}

% where, 
% $L$ - Loss \\
% $N_c$ - Number of initial contours\\
% $A_{min}$ -Minimum area difference (in pixels) between convex hull and ellipse fitting\\
% $C$ - Constant penalty for no contour determination (1000, empirically determined for 640×480)\\

%%%A%% explanation of loss formula based on the 3 factors

\begin{figure}[htb]
    \centering
    \includegraphics[width=0.48\textwidth]{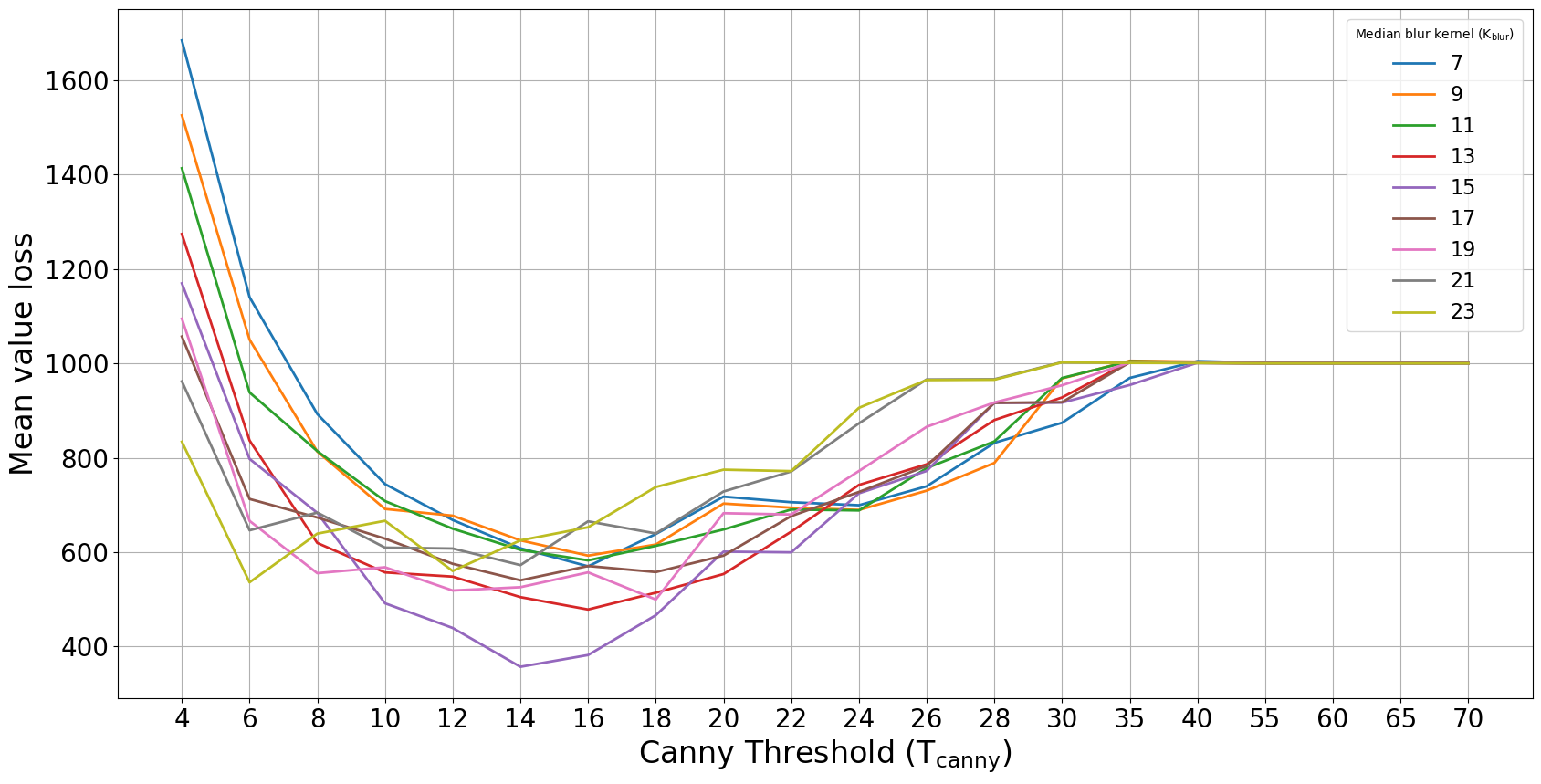}
\caption{Confidence on parameter selection ($T_{canny}$ \& $K_{blur}$) for Subject 04 from our dataset based on the loss formula (eq \ref{chp:evaluation:loss_formula}). The parameters $T_{canny} = 14$ \& $K_{blur} = 15$ display an overall better detection rate.}
\Description{A line plot displaying the loss parameter for different selection of canny and blur value. The plot has a shape of a valley, with the lowest loss value for canny 14 and blur 15.}
    \label{fig:parameter_estimation}
\end{figure}

The loss formula is a confidence measure for the ellipse detection strategy, and gives an estimation on the goodness of the selected parameters. The components of loss criteria include:

$N_c:$ Number of initial contours, is high in case there are more edges in the input image. This happens due to a lower $T_{canny}$ value (see fig \ref{fig:parameter_estimation}), due to which the calculated loss ($L$) is high. Therefore, it is computationally more expensive for our embedded system, and thus it is penalized. For a higher median blur kernel size ($K_{blur}$), $N_c$ is lower in comparison to a low $K_{blur}$ value. 

$C:$ Constant for no contour determination. If no contours/edges are identified in the initial image i.e. $N_c = 0$, there is no possibility to look for a contour that could correspond to pupil. This occurs due to a high $T_{canny}$ value, therefore, the calculated loss value ($L$) is a constant (\emph{C}) penality of 1000. This penality is empirically determined for the resolution 480x640 px. 

$A_{min}:$ Area difference between convex hull and fitted ellipse, is a measure of goodness in the fitted ellipse and convex hull. 
If this measure is high, there is a significant area difference between them, therefore the choice of $T_{canny}$ \& $K_{blur}$ is not effective. If the area enclosed by the convex hull and the area of the fitted ellipse are close to each other, ($A_{min}$), becomes minimum. In case of evaluation of $A_{min}$, there are contours present in the initial image, i.e. $C$ is evaluated as zero.

%These contours are filtered from the initial contours based on area and circularity.  There could be cases where multiple contour edges are detected as prospective pupil, due to which the contour with the minimum difference in area is returned, and it corresponds to the pupil. The parameter, \emph{Amin}, gives an estimate of the fitting of ellipse based on blurring (\emph{K}) and canny threshold value (\emph{Th}).

As can be seen in figure \ref{fig:parameter_estimation}, the value $T_i = 14$ and $K_i = 15$ are the best parameters for estimating pupil in \emph{subject 04}, image resolution (480x640 px). Using this method, we can visualize the selection of parameters that could perform well in the pupil detection pipeline. It is recommended to find the best parameters ($T_{canny}$ \& $K_{blur}$) for each subject before performing pupil detection.

% Figure \ref{fig:parameter_estimation} shows a visual representation of the loss parameter for one participant (\emph{subject 04}) from our dataset. This loss curve is generated by performing the parameter estimation procedure on all the images in the directory. Mean ($\mu$) and standard ($\sigma$) deviation value is calculated for all these loss values and stored in a dictionary. Later, with another set of \emph{Th2} and \emph{K2} parameters, parameter estimation is carried out, for all the images in the working directory. Once all the parameter combinations in this list are used, the final estimation list of $\mu$ and $\sigma$ values are stored in a file to be used to evaluate the best set of parameters. 

%%%%%A%%% removing std would result in the picture to not consume more space

%Latency and memory requirement for pupil detection
\subsection{Runtime \& Memory Consumption} \label{sec:runtime_and_memory_consumption}

Embedded devices are limited in terms of memory and compute resources. It is therefore important to make sure that our programs are not stalling the device by consuming a lot of resources. Our embedded compute device i.e. Raspberry Pi 4B @ 1.5 GHz, 4GB RAM, has enough compute resources to perform our low weight detection method. But performing any graphic intensive processes, could lead to stalling or even crashing of some programs due to low memory. The pipeline for detection method is made using Python and the experimentation was performed on a GUI based operating system. The pupil detection pipeline (\emph{python3} kernel) uses around 100MB of memory and 30 - 35\% of the CPU resources to perform detection on 480x640 px initial image, and 90MB memory with 28 - 30\% CPU resources for 240x320 px. Both of the detections were performed on 30 \emph{fps}. This can be further improved as the pupil camera supports 90 \emph{fps}.

\begin{table*}[ht]
    \centering
    \caption{Average resource consumption during real time ellipse detection at 30$fps$. The memory consumed, and CPU usage varied depending on the type of operation being performed e.g. writing pupil centre locations to a file or displaying a GUI image for pupil detection. These calculations were made with no GUI display and file operations.}
    \begin{tabular}{cccc} 
     \toprule
     \textbf{Experiment} & \textbf{Size of initial image (px)} & \textbf{Memory Consumed (\emph{python3} kernel)} & \textbf{CPU}  \\ [0.5ex] 
     \midrule
     Ellipse detection for 1000 frames & 240x320 & 80-90 MB & 28 - 30\% \\      
     Ellipse detection for 1000 frames & 480x640 & 100 MB &  30 - 35\% \\
    \bottomrule
    \end{tabular}
    \label{table:memory_consumed_ellipse}
\end{table*}

%%%%%Not importent 
% As mentioned above, the OS consumes certain portion of memory for its operation. The detection method consumes additional memory for its processes, and the breakdown of memory consumption during real time ellipse detection method at 30 \emph{fps} is shown in table \ref{table:memory_consumed_ellipse}. As can be seen, the memory consumed for the higher image size (480x640) is more in comparison to the lower initial image size (240x320). This estimation for memory consumed is important because when we add additional features like gaze estimation to our eye tracking system, it could increase the memory consumption even further. We always need to make sure that our programs do not stop the ongoing processes of the operating system.

The time required for each process in the detection pipeline is measured using software timestamps. Python's standard library was used for calculating process timing using a monotonic CPU process timer. The time measurement calculation is performed for 100 frames, where the participant wore the hardware while the detection method is running. The time estimation is made for both image resolutions, 240x320 and 480x640 px. The parameters used for both the resolutions are mentioned in table \ref{table:parameter_time_estimation}.

\begin{table}[ht]
    \centering
    \caption{Parameter selection for computing the time requirement for pupil detection. The table below contains the selection of parameters for both resolution settings of our dataset, 480x640~px. and 240x320~px.}
    \Description{A table displaying the parameter selection for timing estimation of the pupil detector.}
    \begin{tabular}{c c c} 
     \toprule
     \textbf{Parameter} & \textbf{480x640 px} & \textbf{240x320 px} \\ [0.5ex] 
     \midrule
     Canny Threshold ($T_{canny}$) & 24 & 30\\ 
     
     Median Blur Kernel Size ($K_{blur}$) & 23 & 7\\
     
     Max size pupil & 2000 px & 300 px\\
     
     Min size pupil &  1000 px & 100 px\\
     \bottomrule
    \end{tabular}

    \label{table:parameter_time_estimation}
\end{table}

% There is a difference in the parameters used for both the resolutions, and the geometrical size change of the pupil plays a role in it. That is why, the maximum and minimum pupil sizes are changed for different resolutions. These parameters are essential in segmenting the pupil from the rest of the image.

%\subsection{Results for Timing Pupil Detection}

\begin{figure*}[htb]
    \centering
    \includegraphics[width=0.99\textwidth]{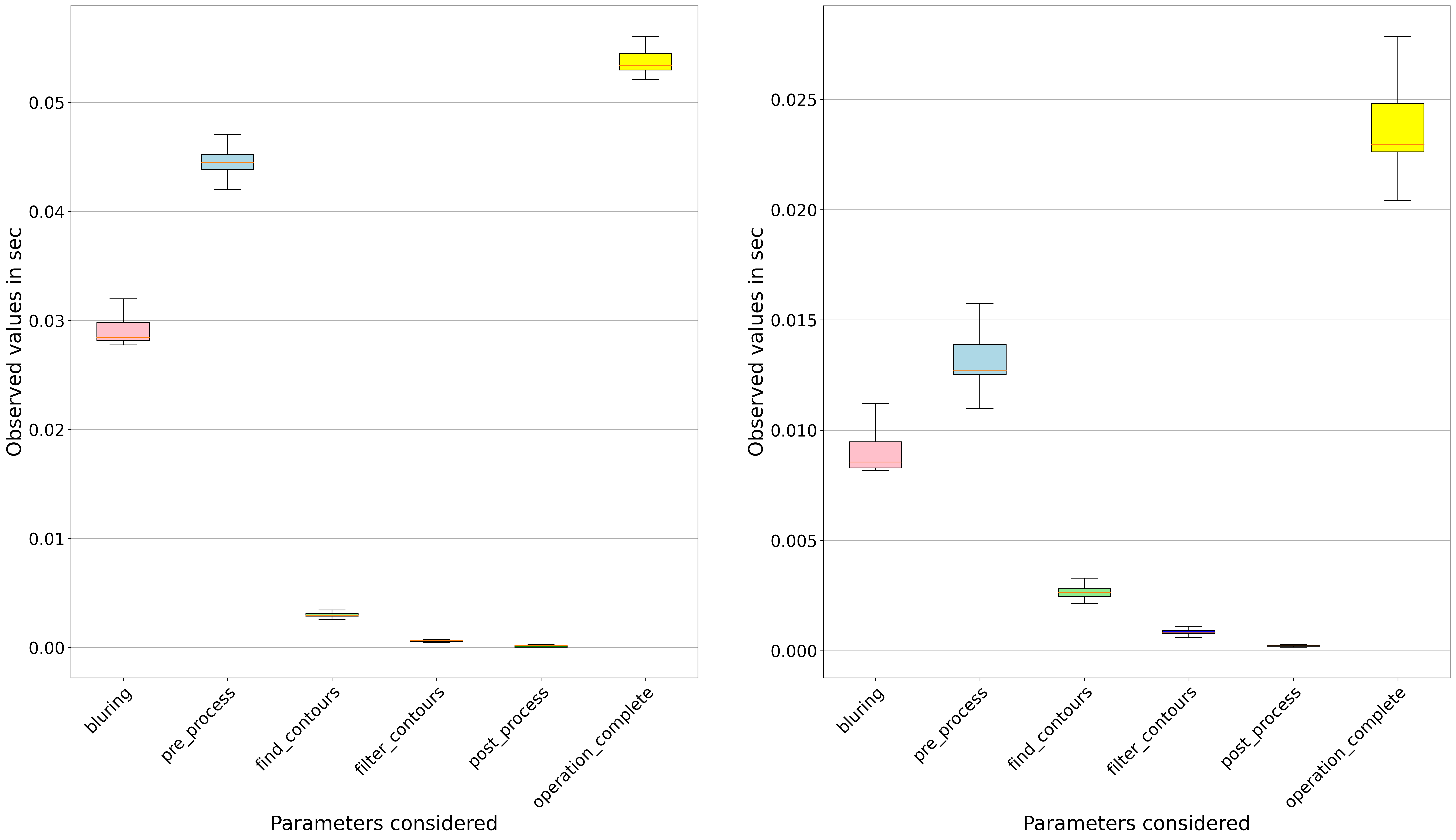}
    \caption{The time required for pupil detection, in linear scale for both resolution settings of 480x640~px (left) and 240x320~px (right). The time measurement is performed on the Raspberry Pi embedded device. Blurring (pre-process) constitutes a significant amount of time, compared to the other steps in the pipeline, accounting for around 50\% in the high and 35\% in the low resolution settings.}
    \Description{A comparison on the timing required to perform pupil detection on the embedded platform for different resolutions.}
    \label{fig:timing_detection_linear}
\end{figure*}

The edge based pupil detection pipeline (fig \ref{fig:pipeline_pupil_detection}), consists of various components like pre-processing, finding of contours, filtering of contours, and post-processing. Figure \ref{fig:timing_detection_linear} shows the breakdown of time required by each component in pupil detection in a box whisker plot on a linear scale. It also shows a comparison of time required to perform pupil detection by different resolutions. The left image is for a higher resolution and the right image is for a lower resolution. As can be seen from figure \ref{fig:timing_detection_linear}, there is a strong difference in finding pupil for both the resolutions while performing the experiment on the hardware. For the lower resolution, the complete operation took 22 ms (median value) for one frame, while it took 50 ms/frame for the higher resolution. The experiment was carried out without any overclocking or acceleration on the Raspberry Pi embedded hardware. For both the experiments, the blurring operation took a significant portion of the total time required for the complete detection process.
% The pre-processing takes into account capturing of image, cropping the image to the required size, gray scaling, median blurring and morphological opening operation. Finding and filtering contours helps in the determination of the contour that corresponds to the pupil, and in post process the location of the pupil is determined and/or saved to a local file.
%The coordinates of the pupil and the size of the pupil is saved to this file. The post-processing step is important because in every gaze estimation set up/pipeline, the coordinates of pupil is used to determine the gaze of the person. 

Our results have also shown that for the lower resolution of 240x320 pixels, the blurring operation has only a minor impact. The detection rate per 100 images for both with and without blurring is the same, i.e., 92 out of 100 frames. This is due to the in-camera downsampling of the image, which leads to an inherent addition of blurring for the lower resolution setting, whereas for the higher resolution of 480x640 pixels, the blurring operation contributes a more significant part of the detection process: Without the blurring operation, the pupil detection performance is lower, with 25-30 frames out of 100 frames, while with the blurring process added in the pipeline, the detection percentage increases to approximately 60-70 frames out of 100.

% Dataset explantion
\subsection{Dataset} \label{sec:dataset}

To study the effect of images taken with our proposed embedded platform, we have collected \textbf{36153} near eye images with our hardware from \textbf{20} participants. We have collected images from people of different ethnicity and age groups and made it as generic as possible to test different pupil detection pipelines. Table \ref{tab:dataset_key_information} contains some key features about the collected data. The procedure of data collection contains two different strategies that allow us to have a diverse set of images, which have been motivated in section \ref{sec:methods}. The dataset from \emph{fixed marker location} can be used as a calibration dataset, containing images of different participants looking at the screen from a fixed distance, or can be used for parameter estimation. The \emph{free movement} images provided enough variations to our dataset and gave us edge cases with blurry images, reflections and close/occluded eye.

\begin{table}[ht]
\centering
\caption{Key features to illustrate the diversity of our collected dataset of near eye infrared images, taken with the embedded platform as described in Section \ref{sec:methods}.}
\begingroup
\setlength{\tabcolsep}{7pt} % Default value: 6pt
\renewcommand{\arraystretch}{1} % Default value: 1
\begin{tabular}{c c c c}
\toprule
\textbf{Parameter}      & \textbf{Value} & \textbf{Parameter}      & \textbf{Value} \\ 
\midrule
Participants   & 20    &   Images        & 36153         \\  
Participant - Male        & 12   &     Nationalities & 9              \\ 
Participant - Female      & 8   &   Average Age & 26                            \\
\bottomrule
\end{tabular}
\endgroup
\label{tab:dataset_key_information}
\end{table}

\begin{figure}[t]
    \centering
    \resizebox{\columnwidth}{!}{%
    \begin{tabular}{ccccc}
    \subcaptionbox{Marker location 1 \label{fig:marker_1}}{\includegraphics[width=0.3\textwidth]{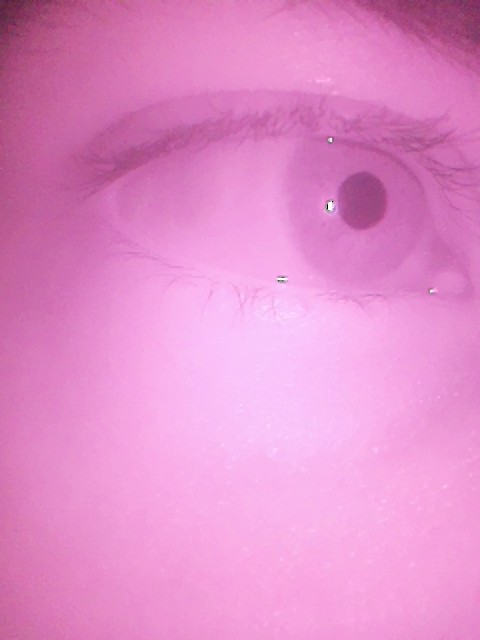}} &
    \subcaptionbox{Marker location 2 \label{fig:marker_2}}{\includegraphics[width=0.3\textwidth]{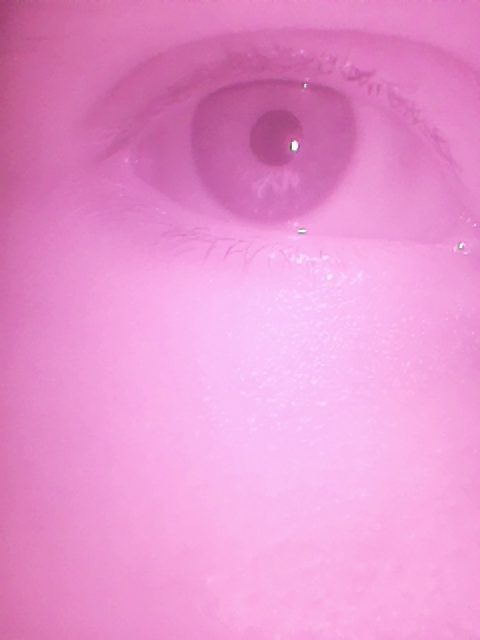}} &
    \subcaptionbox{Marker location 3 \label{fig:marker_3}}{\includegraphics[width=0.3\textwidth]{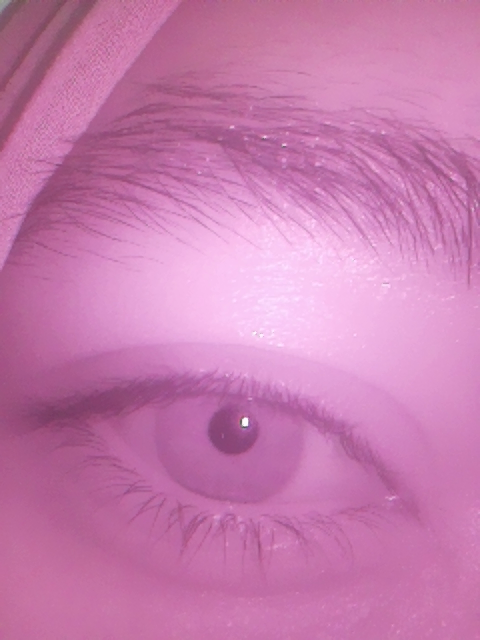}} &
    \subcaptionbox{Marker location 4 \label{fig:marker_4}}{\includegraphics[width=0.3\textwidth]{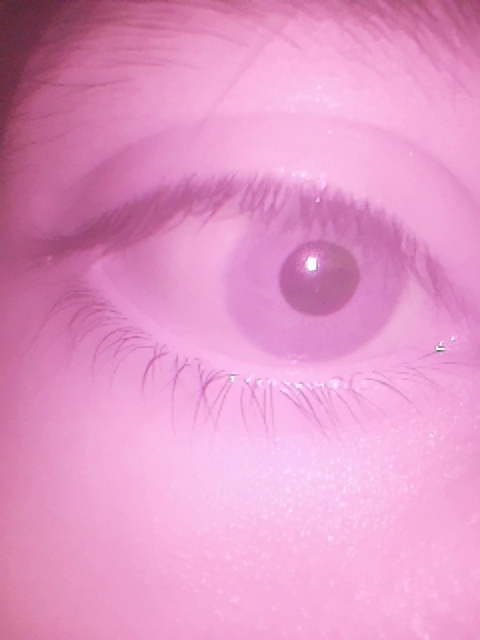}} &
    \subcaptionbox{Marker location 5 \label{fig:marker_5}}{\includegraphics[width=0.3\textwidth]{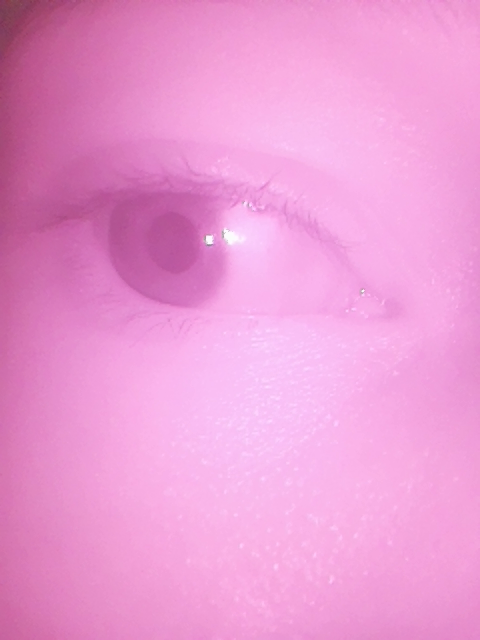}}
    \end{tabular}%
    }
    \Description{A collage of 5 images of diffrent participant's near eye image in infrared lighting. The participants are maintaining gaze on fixed marker locations.}
    \caption{Dataset images from the fixed marker location strategy with resolution 480x640 px. The participants are fixating on different marker locations, due to which there are no blinking/closed eye images. \vspace{-4mm}}
    \label{fig:evaluation_marker_dataset}
\end{figure}

\begin{figure}[t]
    \centering
    \resizebox{\columnwidth}{!}{%
    \begin{tabular}{ccccc}
    \subcaptionbox{ \label{fig:free_movement_1}}{\includegraphics[width=0.3\textwidth]{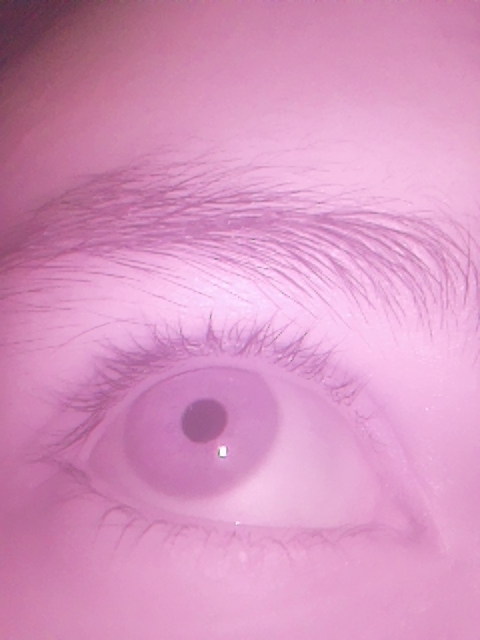}} &
    \subcaptionbox{ \label{fig:free_movement_2}}{\includegraphics[width=0.3\textwidth]{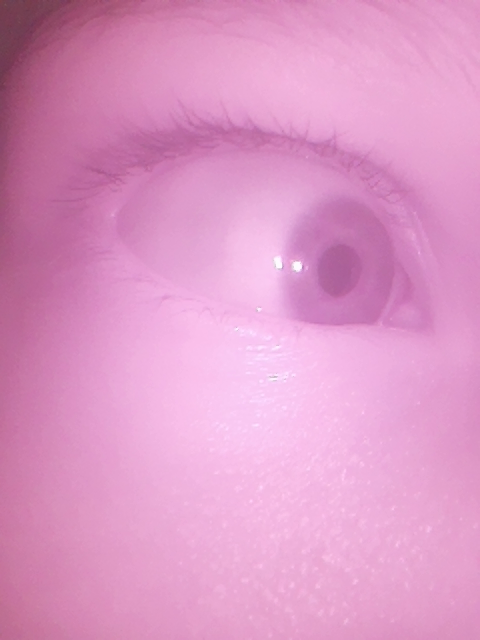}} &
    \subcaptionbox{ \label{fig:free_movement_3}}{\includegraphics[width=0.3\textwidth]{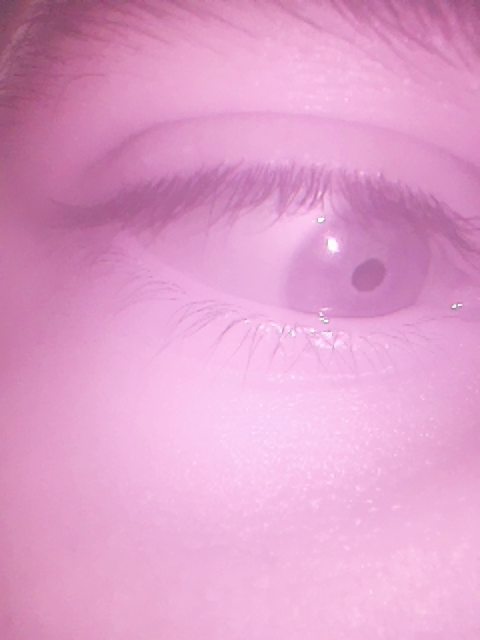}} &
    \subcaptionbox{ \label{fig:free_movement_4}}{\includegraphics[width=0.3\textwidth]{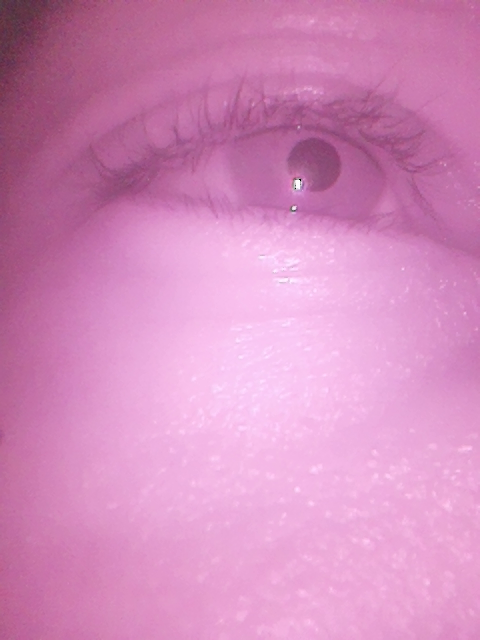}} &
    \subcaptionbox{ \label{fig:free_movement_5}}{\includegraphics[width=0.3\textwidth]{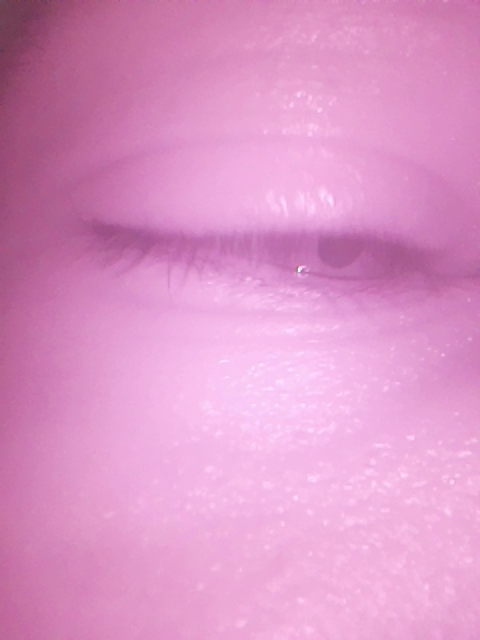}} \\
    \subcaptionbox{
    \label{fig:free_movement_6}}{\includegraphics[width=0.3\textwidth]{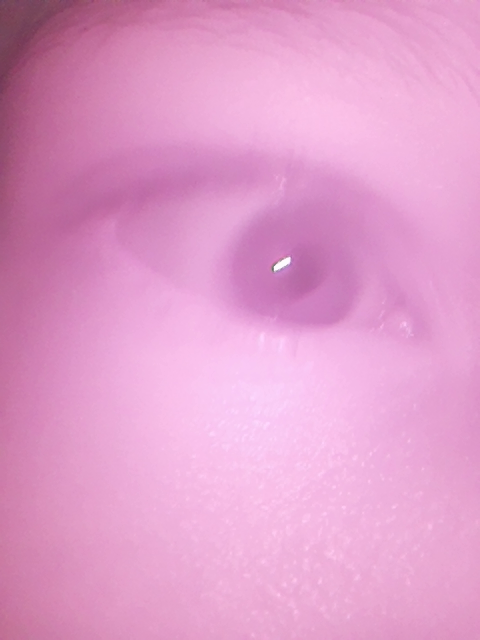}} &
    \subcaptionbox{ \label{fig:free_movement_7}}{\includegraphics[width=0.3\textwidth]{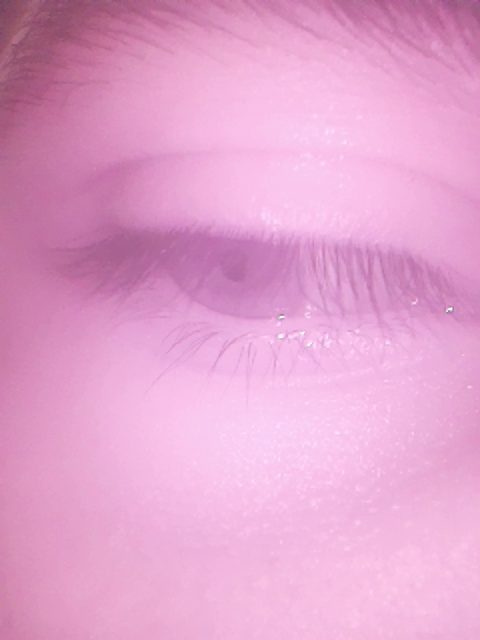}} &
    \subcaptionbox{ \label{fig:free_movement_8}}{\includegraphics[width=0.3\textwidth]{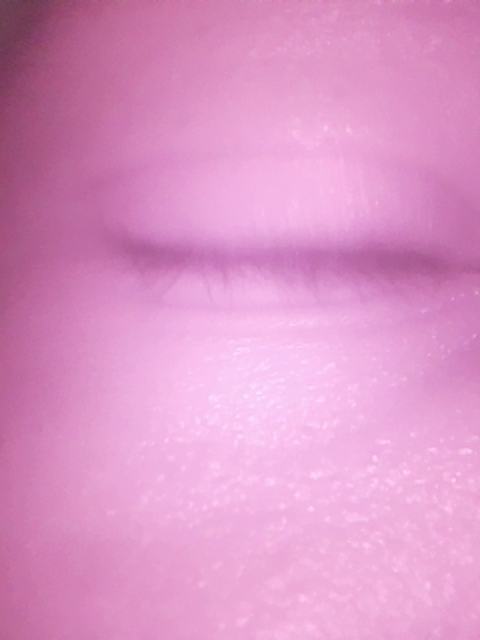}} &
    \subcaptionbox{ \label{fig:free_movement_9}}{\includegraphics[width=0.3\textwidth]{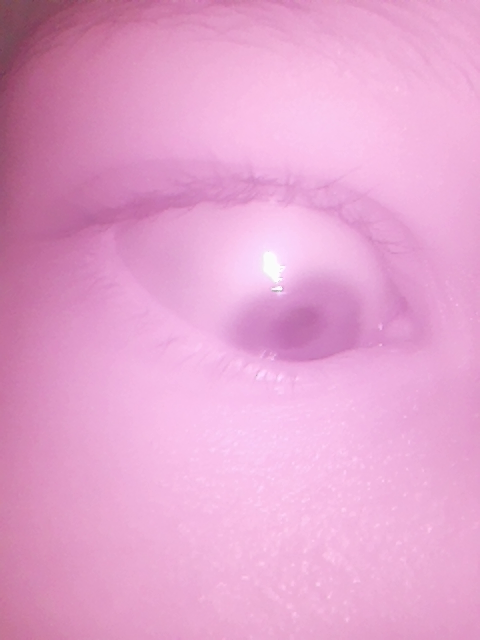}} &
    \subcaptionbox{ \label{fig:free_movement_10}}{\includegraphics[width=0.3\textwidth]{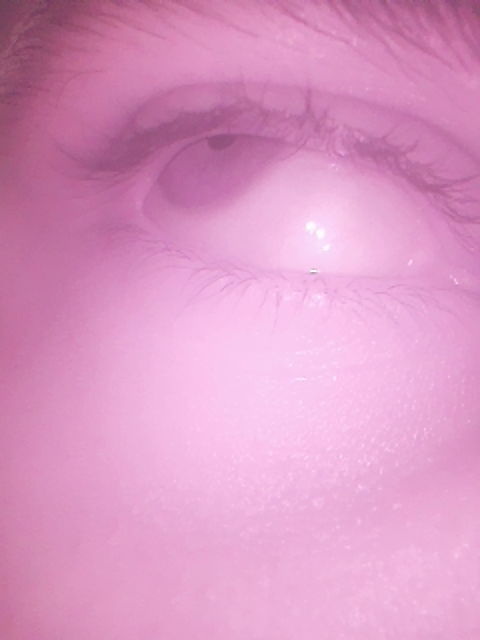}}
    \end{tabular}%
    }
    \Description{A collage of near eye images representing variety of images contained within the proposed dataset for the free movement experiment.}
    \caption{Free movement experiment @ 480x640 px, it contains edge cases as blurry images (fig \ref{fig:free_movement_5}, \ref{fig:free_movement_9}), occluded eyes (fig \ref{fig:free_movement_5}, \ref{fig:free_movement_10}), closed eyes and gaze fixation in various directions (fig \ref{fig:free_movement_1}, \ref{fig:free_movement_3}, \ref{fig:free_movement_7}). \vspace{-4mm}}
    \label{fig:evaluation_free_movement}
\end{figure}

Figure \ref{fig:evaluation_marker_dataset} displays several example images from the fixed marker experiment of the participants, where the participants' gaze is towards the set markers. Due to this setup, there are almost no images where the participant is either blinking or rolling their eye, thereby providing images that can be used as a calibration method for the detector. The marker location dataset contains images in two different resolutions, 480x640 px and 240x320 px. Additionally, these images were also used to identify the parameters for pupil detection, since the pupil is always clearly visible in the frames. On the other hand, figure \ref{fig:evaluation_free_movement} contains some example images from the free movement experiment. Figures \ref{fig:free_movement_1} - \ref{fig:free_movement_4} show images of participants looking in random directions. There are multiple more difficult edge case images within the dataset which would make the detection process challenging. Figures \ref{fig:free_movement_5} - \ref{fig:free_movement_10} show particular examples of some these edge cases. These include blurry images in figure \ref{fig:free_movement_6}, occluded images in figures \ref{fig:free_movement_5} and \ref{fig:free_movement_8}) as well as reflections in figure \ref{fig:free_movement_9}. The dataset contains free movement images in both resolutions of 640x480~px and 320x240~px.

%%%%%%%%%%%%%%%%%%%%%%%%%%%%%%%%%%%%%%%%%%%%%%%%%%%%
\section{Conclusions}\label{sec:conclusions}
%%%%%%%%%%%%%%%%%%%%%%%%%%%%%%%%%%%%%%%%%%%%%%%%%%%%

In this paper, we first presented a real-time and open-source eye tracker setup that allows reproducible pupil detection studies. The hardware design centres around a custom 3D head mount design, which supports a miniature infrared-sensitive eye (Pi Zero) camera, infrared illumination, and a (Logitech C615) world camera. These components are controlled by a single (Raspberry Pi Model 4B) embedded system. 
Using this platform, we contribute with an open-source data pipeline for pupil detection for the eye camera. Based on ellipse fitting and edge analysis, we present a method that can perform real-time pupil detection on the embedded system. 
Furthermore, we have contributed a complementary dataset of near eye images of more than 35000 frames, from 20 participants at 30\emph{fps} for our particular hardware system.
In subsequent experiments, we show that the pupil detection pipeline has a cumulative accuracy of 38.21\% with less than 5px error, 43.6\% with less than 10px error, and an overall average accuracy of 51.9\% at 5.3368 px on the Labelled Pupils in the Wild (LPW) dataset \cite{DBLP:journals/corr/TonsenZSB15}. 

The hardware design files, the detection pipeline source code, and the complementary dataset presented in this paper have been made publicly available to support furthering this research by others and replication of our results. These can be downloaded at: \newline \url{https://github.com/ankurrajw/Pi-Pupil-Detection}.

\begin{acks}
We thank all study participants for their contribution to this work.
\end{acks}

\bibliographystyle{ACM-Reference-Format}
\bibliography{main}

\end{document}